# Oxygen vacancies modulated VO$_2$ for neurons and Spiking Neural Network construction

Liang Li[1], Ting Zhou[1], Tong Liu[1], Zhiwei Liu[2,3,4], Yaping Li[5], Shuo Wu[5], Shanguang Zhao[1], Jinglin Zhu[1], Meiling Liu[1], Zhihan Lin[1], Bowen Sun[1], Jianjun Li[1], Fangwen Sun[2,3,4], Chongwen Zou[1*]

[1] National Synchrotron Radiation Laboratory, School of Nuclear Science and Technology, University of Science and Technology of China, Hefei, Anhui 230029, P. R. China

[2] CAS Key Laboratory of Quantum Information, University of Science and Technology of China, Hefei 230026, China.

[3] CAS Center for Excellence in Quantum Information and Quantum Physics, University of Science and Technology of China, Hefei 230026, China.

[4] Hefei National Laboratory, University of Science and Technology of China, Hefei 230088, China.

[5] Center for Micro- and Nanoscale Research and Fabrication, University of Science and Technology of China, Hefei 230029, China

*Corresponding Author: czou@ustc.edu.cn

# Abstract

Artificial neuronal devices are the basic building blocks for neuromorphic computing systems, which have been motivated by realistic brain emulation. Aiming for these applications, various device concepts have been proposed to mimic the neuronal dynamics and functions. While till now, the artificial neuron devices with high efficiency, high stability and low power consumption are still far from practical application. Due to the special insulator-metal phase transition, Vanadium Dioxide ($VO_2$) has been considered as an idea candidate for neuronal device fabrication. However, its intrinsic insulating state requires the $VO_2$ neuronal device to be driven under large bias voltage, resulting in high power consumption and low frequency. Thus in the current study, we have addressed this challenge by preparing oxygen vacancies modulated $VO_2$ film($VO_{2-x}$) and fabricating the $VO_{2-x}$ neuronal devices for Spiking Neural Networks (SNNs) construction. Results indicate the neuron devices can be operated under lower voltage with improved processing speed. The proposed $VO_{2-x}$ based back-propagation SNNs (BP-SNNs) system, trained with the MNIST dataset, demonstrates excellent accuracy in image recognition. Our study not only demonstrates the $VO_{2-x}$ based neurons and SNN system for practical application, but also offers an effective way to optimize the future neuromorphic computing systems by defect engineering strategy.

# Introduction

The Moore's Law has dominated the modern information industry for over several decades based on the classical von Neumann computing architecture(*1–4*). While as the growing applications in various areas, such as learning, recognition, optimization, and classification, the bottlenecks of processing speed and power efficiency become more pronounced in the traditional computational framework(*5, 6*).

To address the challenges in the post-Moore era, neuromorphic computing, inspired by the brain's architecture, has been proposed to build a computing system that physically emulates the structure and working mechanism of human brains(7,8). For these computing systems, artificial neuronal devices are the basic building blocks, which exhibit considerable potentials in image processing(*9–11*), robotic control(*12*), and combinatorial optimization problem-solving(*13*), etc. Aiming for these functions, various device concepts have been proposed to mimic the neuronal dynamics and functions, such as two-dimensional heterostructures, memristors, phase-change transistors, and ionic devices(*8, 14–16*). While till now, the artificial neuron devices with high efficiency, high stability and low power consumption are still not able to fulfil the practical large-scale device integration.

As a strongly correlated oxide(*17*), Vanadium Dioxide ($VO_2$) has been considered as an idea candidate for neuronal device fabrication due to the special insulator-metal phase transition properties (*18–22*). For example, researchers have developed $VO_2$-based neuronal devices including edge detection systems(*11*), arrhythmia detection systems(*18*), and Ising solvers(*13*). However, the intrinsic insulating state of $VO_2$ requires the fabricated neuronal device to be driven under large bias voltages, resulting in high power consumption and instability. Thus to further modulate the phase transition property of $VO_2$, and achieve $VO_2$-based neuronal device with high efficiency and low power consumption are highly desirable, especially for the integrated neuromorphic computing systems(*14*).

In the current study, we have introduced oxygen vacancies with different concentrations into $VO_2$ films ($VO_{2-x}$) to modulate the intrinsic insulating state as well as the phase transition properties. Then based on the $VO_{2-x}$ structure, we have fabricating low-voltage driven neuronal devices for Spiking Neural Networks (SNNs) construction. Results indicate the $VO_{2-x}$ neuron devices can be operated with improved processing speeds and low power consumption, showing the advantage of the oxygen vacancies modulation effect. The proposed $VO_{2-x}$ based back-propagation SNNs (BP-SNNs) system also shows excellent accuracy in image recognition after trained with the MNIST dataset, demonstrating great potentials for neuromorphic computing in the future.

## Results

**Oxygen vacancies controlled deposition for epitaxial $VO_{2-x}$ films**

From the perspective of defect engineering, various methods have been employed to modulate the phase transitions of $VO_2$, including ion irradiation(*23*, *24*), probe writing(*25*, *26*), carbon nanotube modification(*27*), photothermal regulation(*28*), and ion doping(*29*, *30*). Among them, introducing oxygen vacancy defects into $VO_2$ lattice during the film growth process is a facile and effective way to change the intrinsic insulator state as well as the phase transition behavior (*31*).

Leveraging the superior capabilities of MBE for fabricating non-equilibrium state thin films, high-quality epitaxial $VO_2$ thin films can be successfully grown by the precise vanadium-to-oxygen ratio control. Thus it is easy to introduce the oxygen vacancies with different concentrations into $VO_2$ crystal(*21*). While it should be noticed that the phase transition behavior of the prepared $VO_2$ epitaxial film is always very sensitive to the growth conditions and other factors, such as film thickness, surface topography, and roughness. Thus in order to rigorously isolate and minimize extraneous influences on the Mott device characteristics, we have conducted the film deposition as shown in Figure 1A. By utilizing the distinct profiles of oxygen and vanadium beams, and by adjusting the substrate's position, we can control the V-O ratio at different film locations over a 2-inch diameter substrate.

The obtained 2-inch $VO_{2-x}$ epitaxial film with engineered gradient of oxygen vacancies (as evidenced by the orange dash line) was shown in the optical image of Figure 1B. For the surface roughness measurements, Figure 1C elucidates the data obtained from Scanning Electron Microscopy (SEM) and Atomic Force Microscopy (AFM) across ten points (spaced at 5mm intervals) along the orange dashed line in Figure 1B. The data demonstrate the morphological consistency of the epitaxial $VO_{2-x}$ films with different oxygen vacancies concentrations, showing the similar thickness (~ 55 nm) and surface roughness (Ra = ~ 0.22 nm). In addition, the XRD results also confirmed the excellent epitaxial qualities of $VO_{2-x}$ film (see Supplementary Information Note 1 for details).

Furthermore, we selected three positions on the 2-inch VO$_{2-x}$ film (labeled by three solid points in Figure 1B) for the in-depth electronic structures tests using synchrotron radiation X-ray absorption spectroscopy (XAS). Figures 1D and 1E illustrated the V L-edge ($L_{-II}$ and $L_{-III}$ edges, the transition from V *2p* to V *3d*) and O K-edge for the XAS measurements, respectively. It was observed that the V L-edge shifted towards low photon-energy direction due to the decreased V valance state, which was mainly associated with the oxygen vacancies induced electron doping effect. In addition, it was clear that the higher oxygen vacancies concentration, the more pronounced V-L edge shift existed as indicated by the arrow in Figure 1D. For the O K-edge spectra (transition from O*1s* to O *2p* state) in Figure 1E, two features noted by t$_{2g}$ and e$_g$ were observed, which directly reflected the unoccupied states. It was observed that the relative intensity ratio of the t$_{2g}$ and e$_g$ peaks decreased gradually, indicating a gradual filling of the d$_{//}$ orbital electron states by the emergence of oxygen vacancies. Clearly these electronic evolution detected by XAS tests confirms the successful achievement of the oxygen vacancy concentration gradient in the prepared VO$_{2-x}$ film (*32–34*).

The electrical property modulated by oxygen vacancies with different concentration is also clearly demonstrated by the resistance-temperature (R-T) measurements in Figures 1F and 1G, illustrating the cooling and heating processes, respectively. These two images reveal the resistance-temperature mapping along the profile (the orange dash line) shown in Figure 1B, showing the increase of oxygen vacancy concentration will induce a noticeable decrease in the film's resistance (From the left to right/-25mm to 25mm along the x-axis) at room temperature. In addition, the sharp resistance change (approximately from $10^4$ to 10 Ohm) is also gradually decreased across the phase transition process as the oxygen vacancies concentration increasing. It is noteworthy that for the much higher concentrations of oxygen vacancies, there is no significant reduction in the critical temperature, which suggests that oxygen vacancies play an important role in localized metallization rather than merely lowering the critical temperature (see Supplementary Information Note 3 for details).

**Oscillatory characteristic of two-terminal VO$_{2-x}$ device**

Since the epitaxial VO$_{2-x}$ thin films with variable oxygen vacancy concentration were prepared, it was able to fabricate the two-terminal VO$_{2-x}$ device (Figure 2A) at the selected position on the 2-inch sample with the preferred oxygen vacancy concentration. The VO$_{2-x}$ device is prepared with two Au (20 nm) / Ti (10 nm) counter electrodes on the surface using standard photolithography technique. The SEM image in Figure 2B shows the microstructure of the practical device, while the Figure 2C shows the enlarged 2 μm gap between the two electrodes, which is indicated by the red dashed box in Figure 2B.

For this two-terminal $VO_{2-x}$ device, applying external voltage will drive its phase transition process and show classical threshold switching characteristics as indicated in Figure 2E. It is clear that once the voltage is higher than the threshold value, a conductive filament will quickly formed between the two electrodes due to the local insulator-metal transition of $VO_2$. Then the voltage applied on the device will decrease greatly, makes the metallic filament disappeared and the two electrodes are disconnected again, leading to a continuous oscillating behavior. In fact, by employing the diamond NV center imaging technique, we are able to directly observe this electrically induced switching behavior from the appearance and dissolution of the conductive filaments between the two electrodes, as shown in Figures 2F and 2D, respectively.

Then, we fabricate five devices, termed as Device 1 to 5, on the 2-inch $VO_{2-x}$ film with preferred positions, which are modulated by oxygen vacancies with the increased concentrations. The room-temperature I-V characterizations of these 5 devices, as presented in Figures 2G to 2K, reveal the inversely relationship between the oxygen vacancy concentration and excitation threshold. It is clear that for the increased oxygen vacancies concentration, the driven voltage is gradually decreased from approximately 6.5V to 1.75V from the Device 1 to Device 5. Moreover, the voltage hysteresis is also decreased, thereby highlighting the significant influence of the oxygen vacancies in modulating the fabricated $VO_{2-x}$ device. Figure 2L displays the cumulative plots of threshold voltages across 1000 repeated cycles, illustrating the remarkably stability of the two-terminal device based on the high quality $VO_{2-x}$ structure(*18*, *35*).

By adding suitable matching circuits for these two-terminal devices, distinctive oscillatory behaviors for Device 1 to Device 5 are observed in Figures 2M to 2Q, which demonstrate pronounced variations both in amplitude and frequency. The amplitude change is mainly attributed to the variations of voltage hysteresis, while the frequency adjustment must consider the changes in device impedance and other factors, as detailed in the circuit analysis provided in Supplementary Information Note 2. Intriguingly, the Device 5 demonstrated almost no oscillation (Figure 2Q), which is probably due to the almost metallic-like state induced by the oxygen vacancies with very high concentration. It is clear that the oscillatory phase diagrams of the $VO_{2-x}$ devices can be pronounced modulated by the incorporation of oxygen vacancies, Supplementary Information Note 3 for more comprehensive details.

**The performance of $VO_{2-x}$ based spiking neurons**

The conceptual framework of Spiking Neural Networks (SNN) is rooted in the principles of biomimetic engineering, which contains two fundamental units: the spiking neurons and synaptic structures. Generally, the synapses will interconnect large amount of neurons to form a network as shown in Figure 3E. In our current study, the $VO_2$-based spiking neuron, with its circuit architecture in Figure 3A, is primarily composed of a relaxation oscillator structure. Supplying a constant voltage ($V_{in}$ = constant) to the system, an oscillatory voltage output ($V_{Osci}$) will be produced due to the negative differential resistance attribute of $VO_2$. Concurrently, a discrete spiking pulse voltage signal ($V_{spike}$) can be detected via a small resistor, as evidenced in Figure 3B.

For this $VO_{2-x}$ based spiking neurons, changing the input voltage ($V_{in}$) parameters can also effectively modulate the spiking pulse generation and frequency. For example, Figure 3C demonstrates the neuron's response to a pulsed input, characterized by a 4μs period and a 2μs pulse width. Intriguingly, the neuron maintained its capability to emit spiking signals, albeit with a temporal divergence from the input pulse, demonstrating its Leak Integrated and Fire (LIF) attributes. In Figure 3D, the spiking neuron's response to $V_{in}$ under a 2ms periodic sinusoidal waveform is also tested. The spiking output displays distinct threshold characteristics and the capacity to emit inverse signals, confirming the symmetric architecture of the neuron structure. From the above tests, it is clear that this $VO_{2-x}$ based spiking neuron device as shown in Figure 3A, is effective to transmit both excitatory and inhibitory signals to spiking pulses for Spiking Neural Network construction.

In fact, the external resistance in the circuit structure (Figure 3A) also play important role for the oscillation behavior and performance of the $VO_{2-x}$ based spiking neurons. Comprehensive insights into this oscillation configuration are available in the Supplementary Information Note 2. Accordingly, selecting the Device 1 as the example, we can plot the Tri-state phase diagram for the fabricated neuron device as the function of external resistance in the circuit and applied voltage (Figure 3F), which clearly indicates the un-firing, oscillating, and firing regions. Figure 3G highlights the $V_{spike}$ outputs corresponding to the un-firing, oscillating, and firing states.

Then based on the VO$_{2-x}$ spiking neurons, we fabricate a 2×2 neuronal network as showing in Figure 3H, which employs a configuration of two inverting adders connected in series. This neuronal network can achieve the accumulation of pulse signals from a preceding neuronal layer. And then through numerical summation and amplification, these signals are transmitted to the subsequent neuronal layer, effectively modulating the firing rate (υ) of these neurons and inducing changes in their excitatory or inhibitory states–evidenced by either an increase or decrease in firing rate, respectively. For example, in the 2×2 neural network here, a scenario is presented where υ(2, 1) > υ(1, 1) = υ(1, 2) > υ(2, 2). The input layer neurons, neuron (1, x), are set to have the same firing rate and transmit signals to the subsequent layer of neurons, neuron (2, x), through synapses with different weights. The results in neuron (2, 1) becoming excited with an increased firing rate, while neuron (2, 2) is inhibited with a reduced firing rate. The whole neuromorphic computing process and the anticipated functional outcomes are listed in Figure 3I.

To further explore the Oxygen vacancies modulated VO$_{2-x}$ spiking neuron devices, we have compared the performance for Devices 1 to 5. Firstly, from the Tri-state phase diagrams for different devices (as shown in Figure 3F or Figure S10), we can obtain the diagrams' triple point for each device in Figure 3J. It is clear that there is a noticeable shift for the triple point towards lower matching resistances and operational voltages as the increased oxygen vacancies concentration. As elaborated in Supplementary Information Note 3, this shift indicates that the VO$_{2-x}$ device with higher oxygen concentration can operate at reduced input voltages and lower power consumption, thus benefiting the integrated circuit systems (36).

In addition, for the spiking neuron device, the operational frequency is also a crucial parameter for computational speed. The relationship between operating voltage and spiking pulse frequency for each device is also plotted in Figure 3K. It is observed that the frequency will greatly increase as the increasing operating voltage. It should be noticed that an increase in oxygen vacancies from Device 1 to 4 will lead to a significant reduction in operating voltage, while maintaining the same frequency range. Accordingly, we can estimate the power consumption for each VO$_{2-x}$ neuron device of a neuron under different working frequency in Figure 3L. It can be seen that the introduction of oxygen vacancies can substantially decrease the power requirements of the neuron device.

**The construction of Spiking Neural Network for image recognition**

Based on the above achieved neurons and synaptic structures, we are able to construct Spiking Neural Networks (SNN) through simple circuit connections. Figure 4A shows the framework of the proposed SNNs, which contains a three-layer structure alongside its back-propagation algorithmic flowchart. The backward propagation process normally use a Mean Squared Error (MSE) function as the loss function, which is inspired by conventional back-propagation neural network methodologies (More details in Supplementary Information Note 5). According to this BP-SNNs framework, we have proposed a SNN structure in Figure 4C, which contains 784 (28*28) input layer neurons, 128 neurons within the hidden layer, and 10 neurons in the output layer. Each neuron in the output layer is specifically aligned to represent the numerals '0' through '9'.

Thus in the application of image recognition, the network adopts rate coding mechanisms, which is lying on with the Modified National Institute of Standards and Technology (MNIST) database. Firstly the peripheral circuitry is employed to normalize the pixel intensities and convert them into corresponding voltage signals for the input layer neurons. Then these signals are transformed into spiking pulses with corresponding frequencies, obtaining different weights through the synaptic structures in the SNNs. The end of this forward propagation will be observed in the output layer, where the neuron with the highest rate of excitation prefers to the recognized number.

For example, taking a digit "8" randomly selected from the MNIST dataset, we illustrated the network's operational details during the classification process. Figure 4B presents the intensity distribution mapping for the digit "8", converting each pixel's data corresponding 50 μs pulse signal inputs for the 784 neurons in the input layer. Then these spiking pulse signals will transfer through the synaptic structure to the 128 hidden layer neurons, triggering different excitation and inhibition states in the hidden layer neurons to attain varied frequencies, as shown in the raster plot in Figure 4E. Finally the signals go to the output layer neurons and induce different response as shown in the raster plot in Figure 4F, which indicates that the neuron indexed as 8 achieves maximal excitation and the other neurons shows the inhibition or diminished firing rates. Accordingly, the initial image, the digit "8" in Figure 4B, is recognized by the proposed SNNs (Figure 4C) and lead to the expected output in the statistical histogram in Figure 4D.

In addition, we have trained this proposed SNN architecture using the MNIST dataset, comprising 60,000 entries. This training regimen has resulted in an accuracy rate of approximately 90% after ten epochs, as demonstrated in Figure 4H. The accuracy metrics are derived from the test dataset with 10,000 entries and the test dataset's relevance hot mapping is illustrated in Figure 4G. From this figure, it is clear that a robust correlation exists between the classification outcomes and the anticipated results, confirming the network's excellent classification performance after the MNIST dataset training.

# Conclusion

In this study, we have systematically explored the Oxygen vacancies modulated $VO_2$ film ($VO_{2-x}$) for neurons and Spiking Neural Network for neuromorphic computing. By the deposition of epitaxial $VO_{2-x}$ thin films with varied concentrations of oxygen vacancies, we have revealed that the introduction of oxygen vacancies into $VO_2$ lattice will effectively modulate its electrical properties. This electron doping effect in $VO_{2-x}$ thin film primarily arises from the localized metallization mechanism induced by the local oxygen vacancies insertion. As a result, the $VO_{2-x}$ based neuron devices can operate at lower voltages and consume less power with increased operational frequency, which is favor for the development of $VO_{2-x}$ based Spiking Neural Networks (SNNs).

Furthermore, through the construction of SNNs utilizing $VO_{2-x}$ based neurons and synaptic circuits, coupled with training using the proposed Back-propagation (BP) algorithm, our work shows the potential of these networks in solving complex problems. The experiments and simulations show that the $VO_{2-x}$ based devices can achieve high accuracy rates, exceeding 90% with merely 10 epochs by training with the MNIST dataset. The image recognition by using this SNNs further confirms the practical application in real-world scenarios. Additionally, the reduced sampling times and energy consumption for the oxygen vacancies modulated $VO_{2-x}$ based neurons show some clues for developing more efficient and rapid computing systems.

In conclusion, our research offers a practical strategy to improve the performance of $VO_2$ based neuromorphic computing architectures through oxygen vacancies defect engineering, which opens a possible way for the design, fabrication, and implementation of energy-efficient, high-speed computational models. By realizing the application of $VO_{2-x}$ in neuromorphic systems for the first time, our study will overcome the limitations of current computing paradigms and accelerate the next-generation neuromorphic computing systems.

# Reference

1. U. K. Das, T. K. Bhattacharyya, Opportunities in Device Scaling for 3-nm Node and Beyond: FinFET Versus GAA-FET Versus UFET. *IEEE Trans. Electron Devices* **67**, 2633–2638 (2020).

2. C. E. Leiserson, N. C. Thompson, J. S. Emer, B. C. Kuszmaul, B. W. Lampson, D. Sanchez, T. B. Schardl, There's plenty of room at the Top: What will drive computer performance after Moore's law? *Science* **368**, eaam9744 (2020).

3. H. N. Khan, D. A. Hounshell, E. R. H. Fuchs, Science and research policy at the end of Moore's law. *Nat. Electron.* **1**, 14–21 (2018).

4. J. Shalf, The future of computing beyond Moore's Law. *Philos. Trans. R. Soc. Math. Phys. Eng. Sci.* **378**, 20190061 (2020).

5. H. Wang, T. Fu, Y. Du, W. Gao, K. Huang, Z. Liu, P. Chandak, S. Liu, P. Van Katwyk, A. Deac, A. Anandkumar, K. Bergen, C. P. Gomes, S. Ho, P. Kohli, J. Lasenby, J. Leskovec, T.-Y. Liu, A. Manrai, D. Marks, B. Ramsundar, L. Song, J. Sun, J. Tang, P. Veličković, M. Welling, L. Zhang, C. W. Coley, Y. Bengio, M. Zitnik, Scientific discovery in the age of artificial intelligence. *Nature* **620**, 47–60 (2023).

6. S. Teng, X. Hu, P. Deng, B. Li, Y. Li, Y. Ai, D. Yang, L. Li, Z. Xuanyuan, F. Zhu, L. Chen, Motion Planning for Autonomous Driving: The State of the Art and Future Perspectives. *IEEE Trans. Intell. Veh.* **8**, 3692–3711 (2023).

7. IEEE International Roadmap for Devices and Systems - IEEE IRDS[TM]. https://irds.ieee.org/.

8. D. Marković, A. Mizrahi, D. Querlioz, J. Grollier, Physics for neuromorphic computing. *Nat. Rev. Phys.* **2**, 499–510 (2020).

9. G. Li, D. Xie, Z. Zhang, Q. Zhou, H. Zhong, H. Ni, J. Wang, E. Guo, M. He, C. Wang, G. Yang, K. Jin, C. Ge, Flexible VO2 Films for In-Sensor Computing with Ultraviolet Light. *Adv. Funct. Mater.* **32**, 2203074 (2022).

10. G. Li, D. Xie, H. Zhong, Z. Zhang, X. Fu, Q. Zhou, Q. Li, H. Ni, J. Wang, E. Guo, M. He, C. Wang, G. Yang, K. Jin, C. Ge, Photo-induced non-volatile VO2 phase transition for neuromorphic ultraviolet sensors. *Nat. Commun.* **13**, 1729 (2022).

11. S. Oh, Y. Shi, J. del Valle, P. Salev, Y. Lu, Z. Huang, Y. Kalcheim, I. K. Schuller, D. Kuzum, Energy-efficient Mott activation neuron for full-hardware implementation of neural networks. *Nat. Nanotechnol.* **16**, 680–687 (2021).

12. S. Dutta, A. Parihar, A. Khanna, J. Gomez, W. Chakraborty, M. Jerry, B. Grisafe, A. Raychowdhury, S. Datta, Programmable coupled oscillators for synchronized locomotion. *Nat. Commun.* **10**, 3299 (2019).

13. S. Dutta, A. Khanna, A. S. Assoa, H. Paik, D. G. Schlom, Z. Toroczkai, A. Raychowdhury, S. Datta, An Ising Hamiltonian solver based on coupled stochastic phase-transition nano-oscillators. *Nat. Electron.* **4**, 502–512 (2021).

14. J. Bian, Z. Cao, P. Zhou, Neuromorphic computing: Devices, hardware, and system application facilitated by two-dimensional materials. *Appl. Phys. Rev.* **8**, 041313 (2021).

15. D. Marković, J. Grollier, Quantum neuromorphic computing. *Appl. Phys. Lett.* **117**, 150501 (2020).

16. Y. Zhang, Q. Zheng, X. Zhu, Z. Yuan, K. Xia, Spintronic devices for neuromorphic computing. *Sci. China Phys. Mech. Astron.* **63**, 277531 (2020).

17. L. L. Fan, Y. F. Wu, C. Si, G. Q. Pan, C. W. Zou, Z. Y. Wu, Synchrotron radiation study of VO2 crystal film epitaxial growth on sapphire substrate with intrinsic multi-domains. *Appl. Phys. Lett.* **102**, 011604 (2013).

18. R. Yuan, P. J. Tiw, L. Cai, Z. Yang, C. Liu, T. Zhang, C. Ge, R. Huang, Y. Yang, A neuromorphic physiological signal processing system based on VO2 memristor for next-generation human-machine interface. *Nat. Commun.* **14**, 3695 (2023).

19. R. Yuan, Q. Duan, P. J. Tiw, G. Li, Z. Xiao, Z. Jing, K. Yang, C. Liu, C. Ge, R. Huang, Y. Yang, A calibratable sensory neuron based on epitaxial VO2 for spike-based neuromorphic multisensory system. *Nat. Commun.* **13**, 3973 (2022).

20. Q. Duan, T. Zhang, C. Liu, R. Yuan, G. Li, P. Jun Tiw, K. Yang, C. Ge, Y. Yang, R. Huang, Artificial Multisensory Neurons with Fused Haptic and Temperature Perception for Multimodal In-Sensor Computing. *Adv. Intell. Syst.* **4**, 2200039 (2022).

21. L. L. Fan, S. Chen, Y. F. Wu, F. H. Chen, W. S. Chu, X. Chen, C. W. Zou, Z. Y. Wu, Growth and phase transition characteristics of pure M-phase VO2 epitaxial film prepared by oxide molecular beam epitaxy. *Appl. Phys. Lett.* **103**, 131914 (2013).

22. H. Ren, B. Li, X. Zhou, S. Chen, Y. Li, C. Hu, J. Tian, G. Zhang, Y. Pan, C. Zou, Wafer-size VO2 film prepared by water-vapor oxidant. *Appl. Surf. Sci.* **525**, 146642 (2020).

23. Y. Zhang, Y. Wang, Y. Wu, X. Shu, F. Zhang, H. Peng, S. Shen, N. Ogawa, J. Zhu, P. Yu, Artificially controlled nanoscale chemical reduction in VO2 through electron beam illumination. *Nat. Commun.* **14**, 4012 (2023).

24. C. Hu, L. Li, X. Wen, Y. Chen, B. Li, H. Ren, S. Zhao, C. Zou, Manipulating the anisotropic phase separation in strained VO2 epitaxial films by nanoscale ion-implantation. *Appl. Phys. Lett.* **119**, 121101 (2021).

25. W. Zhang, X. Wu, L. Li, C. Zou, Y. Chen, Fabrication of a VO2-Based Tunable Metasurface by Electric-Field Scanning Probe Lithography with Precise Depth Control. *ACS Appl. Mater. Interfaces* **15**, 13517–13525 (2023).

26. F. Zhang, Y. Zhang, L. Li, X. Mou, H. Peng, S. Shen, M. Wang, K. Xiao, S.-H. Ji, D. Yi, T. Nan, J. Tang, P. Yu, Nanoscale multistate resistive switching in WO3 through scanning probe induced proton evolution. *Nat. Commun.* **14**, 3950 (2023).

27. S. M. Bohaichuk, S. Kumar, G. Pitner, C. J. McClellan, J. Jeong, M. G. Samant, H.-. S. P. Wong, S. S. P. Parkin, R. S. Williams, E. Pop, Fast Spiking of a Mott $VO_2$ –Carbon Nanotube Composite Device. *Nano Lett.* **19**, 6751–6755 (2019).

28. C. Feng, B.-W. Li, Y. Dong, X.-D. Chen, Y. Zheng, Z.-H. Wang, H.-B. Lin, W. Jiang, S.-C. Zhang, C.-W. Zou, G.-C. Guo, F.-W. Sun, Quantum imaging of the reconfigurable VO2 synaptic electronics for neuromorphic computing. *Sci. Adv.* **9**, eadg9376 (2023).

29. Y. Chen, Z. Wang, S. Chen, H. Ren, L. Wang, G. Zhang, Y. Lu, J. Jiang, C. Zou, Y. Luo, Non-catalytic hydrogenation of VO2 in acid solution. *Nat. Commun.* **9**, 818 (2018).

30. B. Li, L. Li, H. Ren, Y. Lu, F. Peng, Y. Chen, C. Hu, G. Zhang, C. Zou, Photoassisted Electron–Ion Synergic Doping Induced Phase Transition of n-$VO_2$/p-GaN Thin-Film Heterojunction. *ACS Appl. Mater. Interfaces* **13**, 43562–43572 (2021).

31. B. Li, M. Hu, H. Ren, C. Hu, L. Li, G. Zhang, J. Jiang, C. Zou, Atomic Origin for Hydrogenation Promoted Bulk Oxygen Vacancies Removal in Vanadium Dioxide. *J. Phys. Chem. Lett.* **11**, 10045–10051 (2020).

32. L. Fan, X. Wang, F. Wang, Q. Zhang, L. Zhu, Q. Meng, B. Wang, Z. Zhang, C. Zou, Revealing the role of oxygen vacancies on the phase transition of VO2 film from the optical-constant measurements. *RSC Adv.* **8**, 19151–19156 (2018).

33. Z. Zhang, F. Zuo, C. Wan, A. Dutta, J. Kim, J. Rensberg, R. Nawrodt, H. H. Park, T. J. Larrabee, X. Guan, Y. Zhou, S. M. Prokes, C. Ronning, V. M. Shalaev, A. Boltasseva, M. A. Kats, S. Ramanathan, Evolution of Metallicity in Vanadium Dioxide by Creation of Oxygen Vacancies. *Phys. Rev. Appl.* **7**, 034008 (2017).

34. Y. Park, H. Sim, M. Jo, G.-Y. Kim, D. Yoon, H. Han, Y. Kim, K. Song, D. Lee, S.-Y. Choi, J. Son, Directional ionic transport across the oxide interface enables low-temperature epitaxy of rutile TiO2. *Nat. Commun.* **11**, 1401 (2020).

35. J. Narayan, V. M. Bhosle, Phase transition and critical issues in structure-property correlations of vanadium oxide. *J. Appl. Phys.* **100**, 103524 (2006).

36. K. Zhu, C. Wen, A. A. Aljarb, F. Xue, X. Xu, V. Tung, X. Zhang, H. N. Alshareef, M. Lanza, The development of integrated circuits based on two-dimensional materials. *Nat. Electron.* **4**, 775–785 (2021).

**Figures**

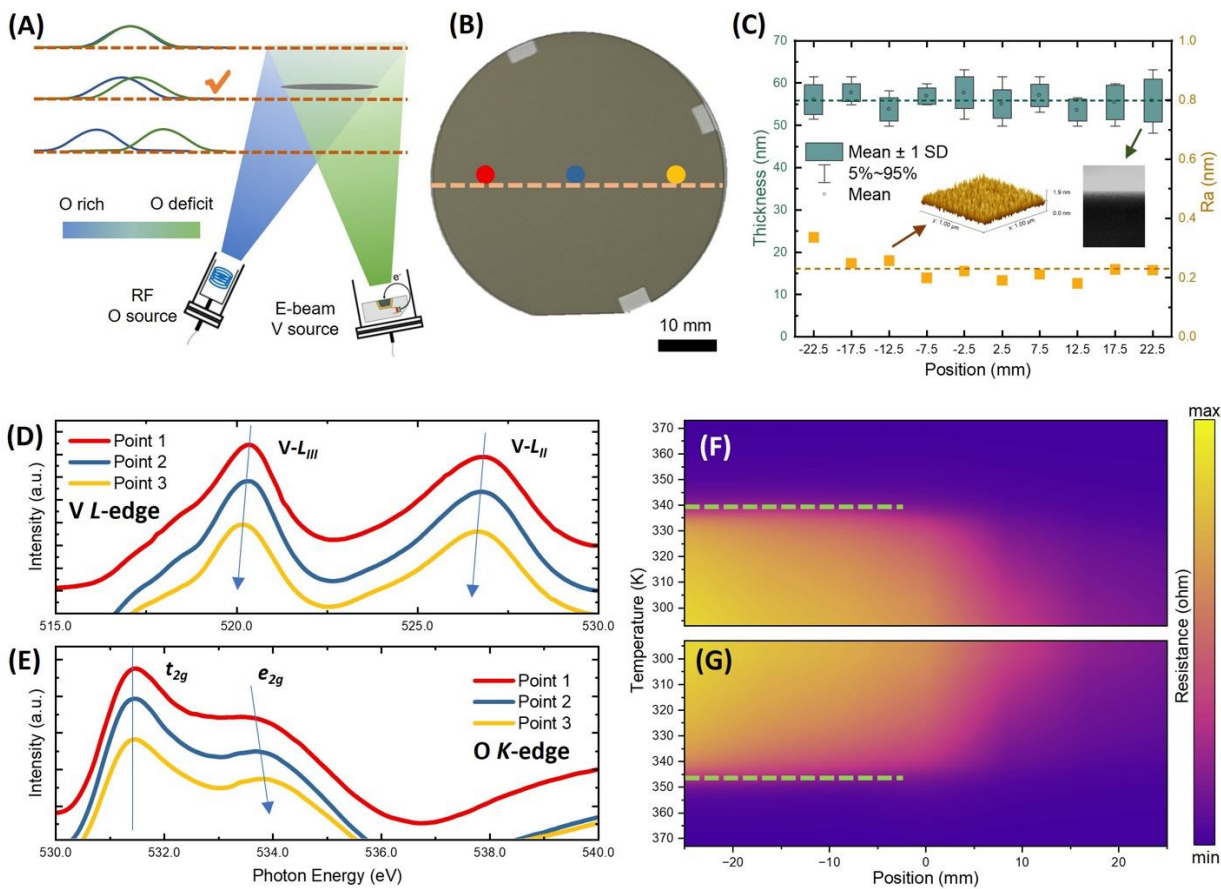

**Figure 1. A**, Schematic illustration of the $VO_{2-x}$ film deposition with the desired gradient of oxygen vacancies via Molecular Beam Epitaxy (MBE). **B**, Optical photograph for the deposited 2-inch $VO_{2-x}$ epitaxial film. The orange dash line shows the profile for the following tests. **C**, Surface roughness measured along the orange dashed line in (B) by Atomic Force Microscopy (AFM), and film thickness measured by Scanning Electron Microscopy (SEM), with sampling points spaced 5 mm apart. The dashed lines represent the obtained average values, which are 0.22 nm for roughness and 55 nm for thickness, respectively. Corresponding AFM topography images and cross-sectional SEM images are available in Supplementary Information Note 1. **D** and **E**, Synchrotron radiation-based X-ray Absorption spectra (XAS) for the three points shown in (B), for the V L-edge and O K-edge, respectively. **F** and **G**, Temperature-dependent resistance-location mapping along the orange dashed line in (B) with cooling and heating processes, respectively.

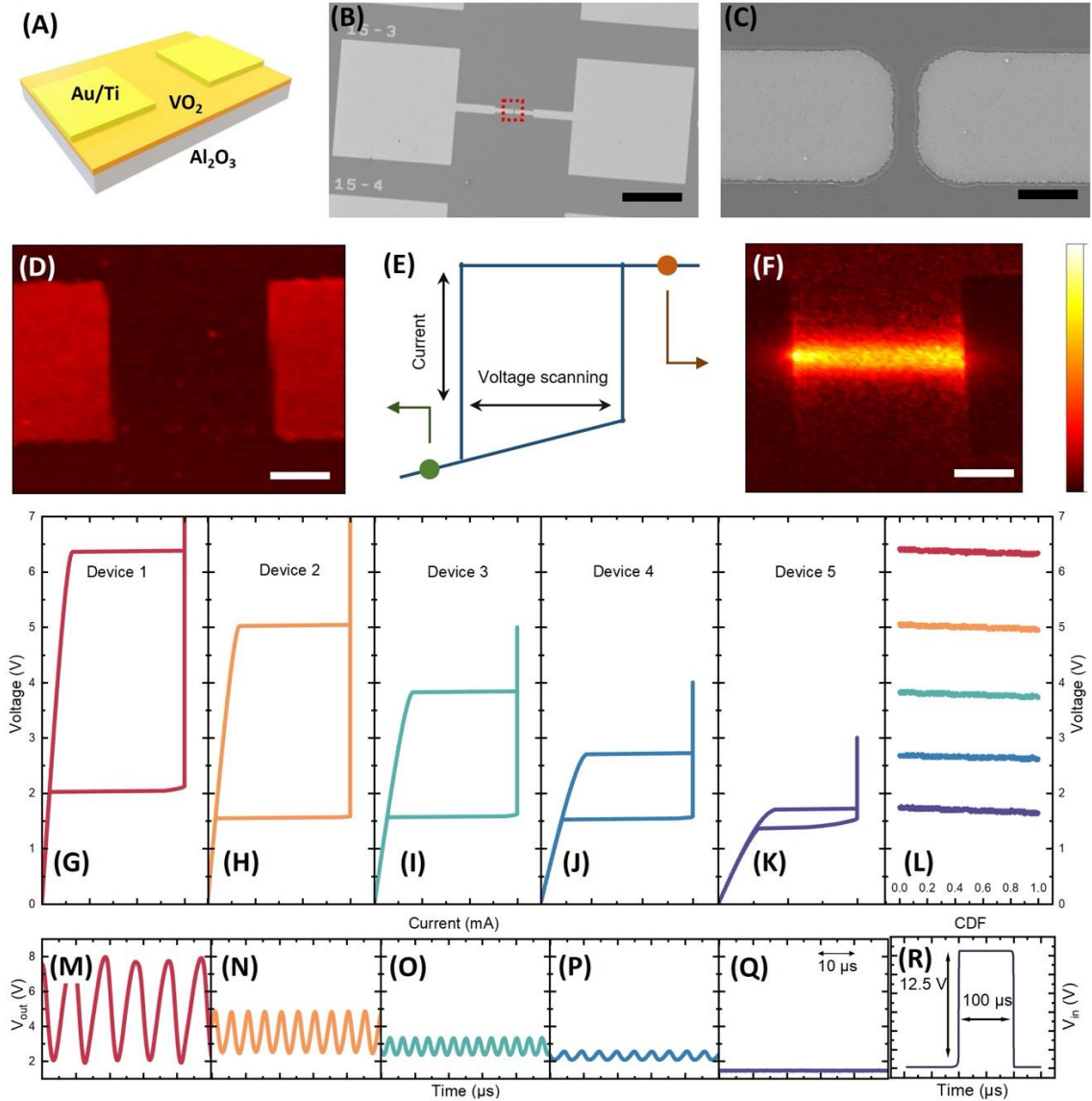

**Figure 2. A**, Schematic diagram of the device structure. **B** and **C**, Scanning Electron Microscopy (SEM) images for the fabricated devices with different magnifications. The scale bars are 100 μm and 5 μm, respectively. **E**, Illustration of the current-voltage (I-V) curve for the two-terminal device during the voltage scanning process. **D** and **F**, The current distribution mapping by NV sensor for the two-terminal device before and after excited by an external voltage. The excited state in (F) clearly shows the conductive filament. The scale bar is 10μm. **G** to **K**, the I-V curves for the Devices 1 to Device 5. **L**, The threshold voltage measurements for over 1000 scanning cycles. **M** to **Q**, The oscillating characteristics for the Devices 1 to Device 5 under the same matching circuit, with a 3 kΩ series resistor and a 1.8 nF parallel capacitor. **R**, A square wave pulse input with a width of 100 μs and an amplitude of 12.5 V.

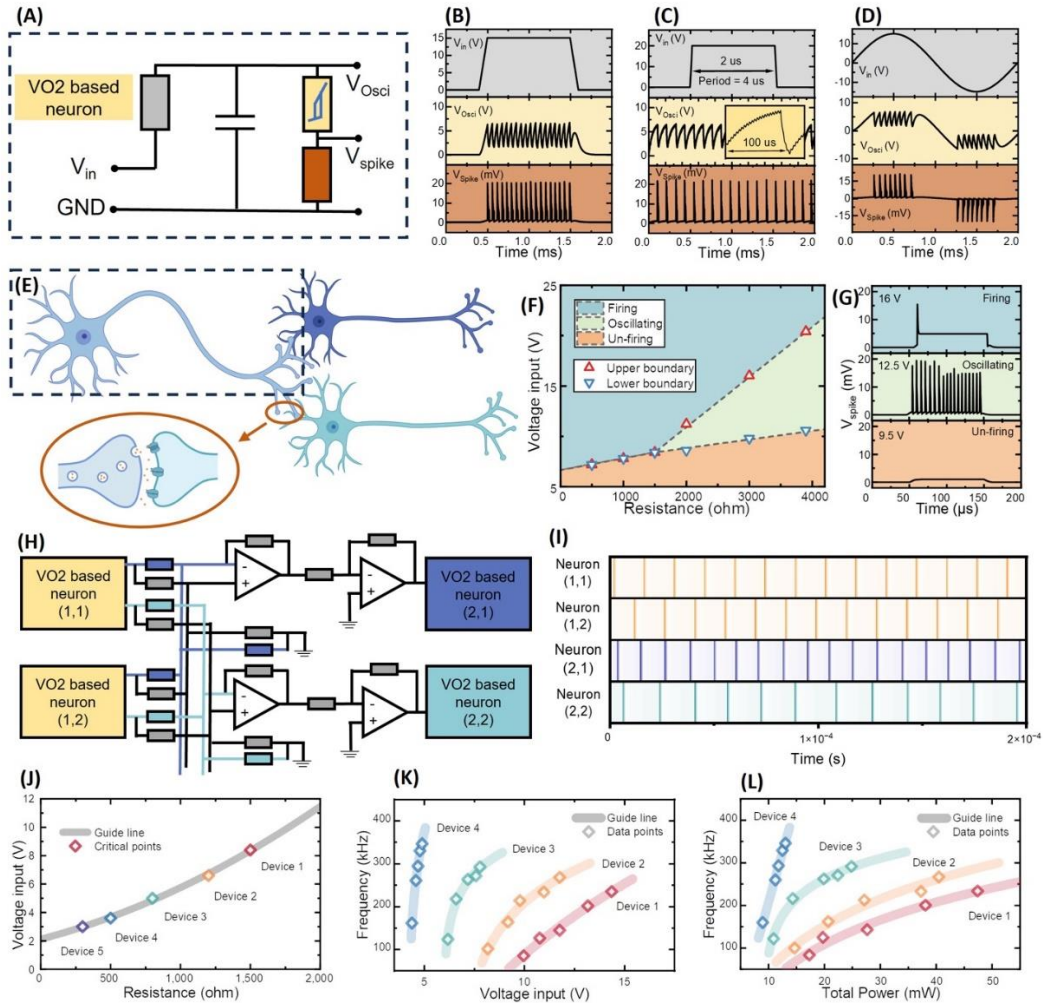

**Figure 3. A**, Schematic diagram of the circuit structure for a $VO_2$-based neuron, which primarily includes a series matching resistor, a parallel capacitor, and a small resistor for spiking signal sampling. **B** to **D**, Output of the designed neuron under different input modes: constant, pulse square wave, and sine wave, respectively. **E**, A scheme for brain-like network including the neurons and synaptic structures. **F**, Tri-state phase diagram of a $VO_2$-based neuron (Device 1). **G**, The typical outputs for the three states in the phase diagram (F). **H**, A scheme for a 2×2 network composed of four neurons. **I**, The Raster plot for the 2×2 neural network, simulated by the spice model of the circuit structure in (H). **J**, Phase diagrams' triple points for Device 1 to 5, extracted from the phase diagrams in Supplementary Information Note 3. **K** and **L**, The Voltage-Frequency curves and Total power-Frequency curves for different devices, respectively.

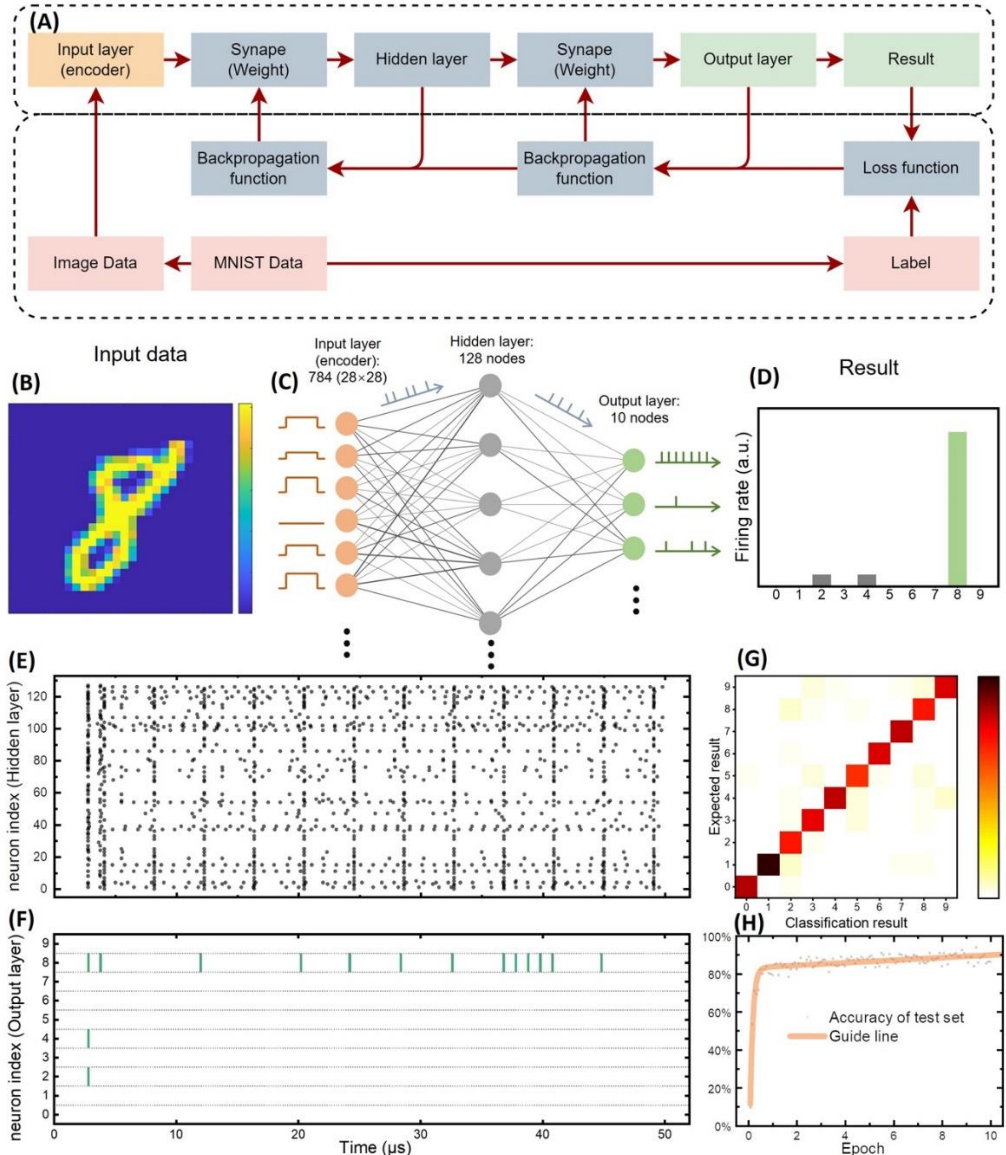

**Figure 4. A**, Scheme for the proposed Back-propagation Spiking Neural Networks (BP-SNNs) architecture. **B**, A randomly selected from the MNIST dataset. **C**, Schematic diagram of the constructed three-layer SNNs structure. **D**, The firing rates histogram from the output layer neurons by using the digit '8' in (B) as the input. **E** and **F**, Raster plots of the hidden layer and output layer, respectively, during network operation over 50 μs. **G**, Confusion matrix of the classification results for the test dataset after 10 epochs, demonstrating the images can be well classified. **H**, The evolution of the test accuracy across training epochs. After only 10 epochs of training, the accuracy can achieve approximately 90%.